\documentclass[10pt,twocolumn,letterpaper]{article}

\usepackage{cvpr}
\usepackage{times}
\usepackage{epsfig}
\usepackage{graphicx}
\usepackage{adjustbox}
\usepackage{amsmath}
\usepackage{amssymb}
\usepackage{multirow}
\usepackage{color}
\usepackage{wasysym}
\usepackage[normalem]{ulem} 
\usepackage{authblk}
\usepackage{enumitem}

\newcommand*\samethanks[1][\value{footnote}]{\footnotemark[#1]}

\usepackage[pagebackref=true,breaklinks=true,letterpaper=true,colorlinks,bookmarks=false]{hyperref}

\cvprfinalcopy 

\def\cvprPaperID{} 

\ifcvprfinal\fi
\begin{document}

\title{Pose-Robust Face Recognition via Deep Residual Equivariant Mapping}


\author{
Kaidi Cao$^{2}$\thanks{indicates shared first authorship} \hspace{9pt} Yu Rong$^{1,2}$\samethanks \hspace{9pt} Cheng Li$^{2}$ \hspace{9pt} Xiaoou Tang$^{1}$ \hspace{8pt}  Chen Change Loy$^{1}$\\
$^{1}$\small{Department of Information Engineering, The Chinese University of Hong Kong}\\
$^{2}$\small{SenseTime Research}\\
{\tt\small \{ry017, ccloy, xtang\}@ie.cuhk.edu.hk \hspace{5pt} \{caokaidi, chengli\}@sensetime.com}
}

\maketitle
\thispagestyle{empty}

\begin{abstract}
Face recognition achieves exceptional success thanks to the emergence of deep learning. However, many contemporary face recognition models still perform relatively poor in processing profile faces compared to frontal faces. A key reason is that the number of frontal and profile training faces are highly imbalanced - there are extensively more frontal training samples compared to profile ones. In addition, it is intrinsically hard to learn a deep representation that is geometrically invariant to large pose variations.
In this study, we hypothesize that there is an inherent mapping between frontal and profile faces, and consequently, their discrepancy in the deep representation space can be bridged by an equivariant mapping. To exploit this mapping, we formulate a novel Deep Residual EquivAriant Mapping (DREAM) block, which is capable of adaptively adding residuals to the input deep representation to transform a profile face representation to a canonical pose that simplifies recognition.
The DREAM block consistently enhances the performance of profile face recognition for many strong deep networks, including ResNet models, without deliberately augmenting training data of profile faces. The block is easy to use, light-weight, and can be implemented with a negligible computational overhead \footnote{Codes and models are available at \url{http://mmlab.ie.cuhk.edu.hk/projects/DREAM/}}. 
\end{abstract}
\section{Introduction}
\label{sec:introduction}

The emergence of deep learning greatly advances the frontier of face recognition~\cite{sun2014deep,taigman2014deepface}. The main focus tends to center around near-frontal faces
while there can be no assurance of view consistency when face recognition is conducted in unconstrained environments.
Although human performance only drops slightly from frontal-frontal to frontal-profile face verification, many existing algorithms can suffer a drop of over 10\%~\cite{sengupta2016frontal}.
Thus, large pose variation remains to be a significant challenge that confronts real-world face recognition.

\begin{figure}[t]
\begin{center}
   \includegraphics[width=1\linewidth]{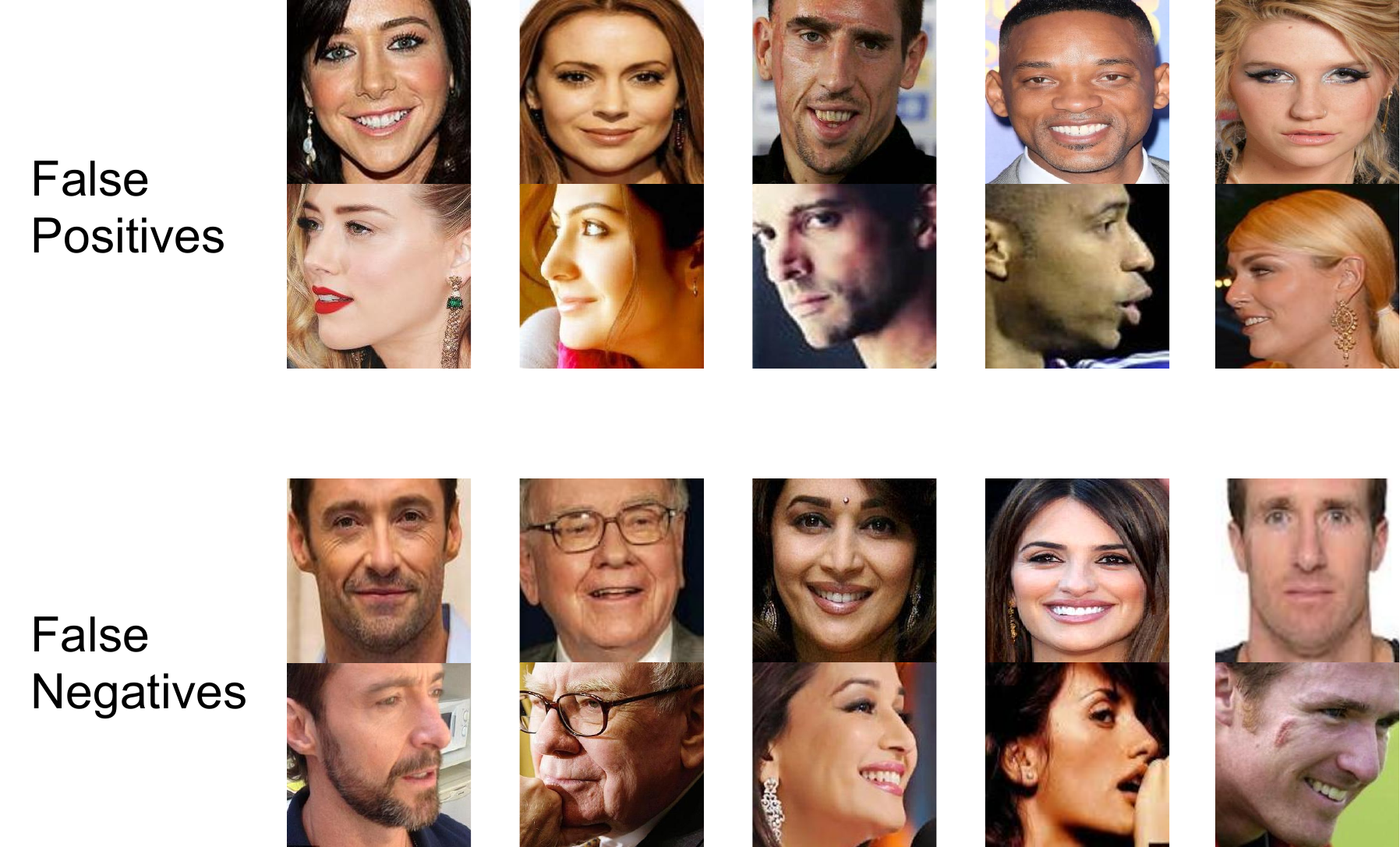}
\end{center}
\vskip -0.4cm
   \caption{A state-of-the-art face recognition model~\cite{wu2016deep} tested on a challenging frontal-profile faces dataset~\cite{sengupta2016frontal}. 
   It is observed that profile faces of different persons are easily to be mismatched (false positives),
   and profile and frontal faces of the same identity may not trigger a match leading to false negatives.}
\label{fig:superROC}
\vspace{-0.25cm}
\vskip -0.3cm
\end{figure}

We provide an example in Figure~\ref{fig:superROC} to show the failure modes of a state-of-the-art face verification model. We trained the same ResNet-18 model as reported in~\cite{wu2016deep}. This model achieves a high accuracy of 99.3\% on the LFW benchmark~\cite{huang2007labeled}. Despite the strong model, it tends to falsely match profile faces of different identities yielding a number of false positives. In addition, the model is also likely to miss frontal and profile faces of the same identity leading to false negatives. 

Why does face recognition work poorly on profile faces? Modern deep learning is heavily data-driven \cite{huang2016learning,guo2016ms}. The generalization power of deep models is usually proportional to the training data size. Given an uneven distribution of profile and frontal faces in the dataset, deeply learned features tend to bias on distinguishing frontal faces rather than profile faces.
When it is infeasible to collect a massive dataset that covers all possible poses with even distribution, researchers have turned to alternative approaches to better handle the recognition of profile faces. 
A large body of methods normalize images to a single frontal pose before recognition, either through elaborated dense 3D facial landmark detection and warping~\cite{taigman2014deepface}, or another deep model (or generative adversarial network) specialized in face frontalization~\cite{tran2017disentangled}. Such methods would add processing burden to the whole system. In addition, face frontalization in the wild, especially with extreme profile faces, is still considered challenging. Often, synthesized `frontal' faces would contain artifacts caused by occlusions and non-rigid expressions. Another potential solution is divide-and-conquer, \ie, training separate models for learning pose-specific identity features~\cite{masi2016pose}. This strategy tends to increase computational cost due to the use of multiple models.
%

\begin{figure}[t]
\begin{center}
\includegraphics[width=1\linewidth]{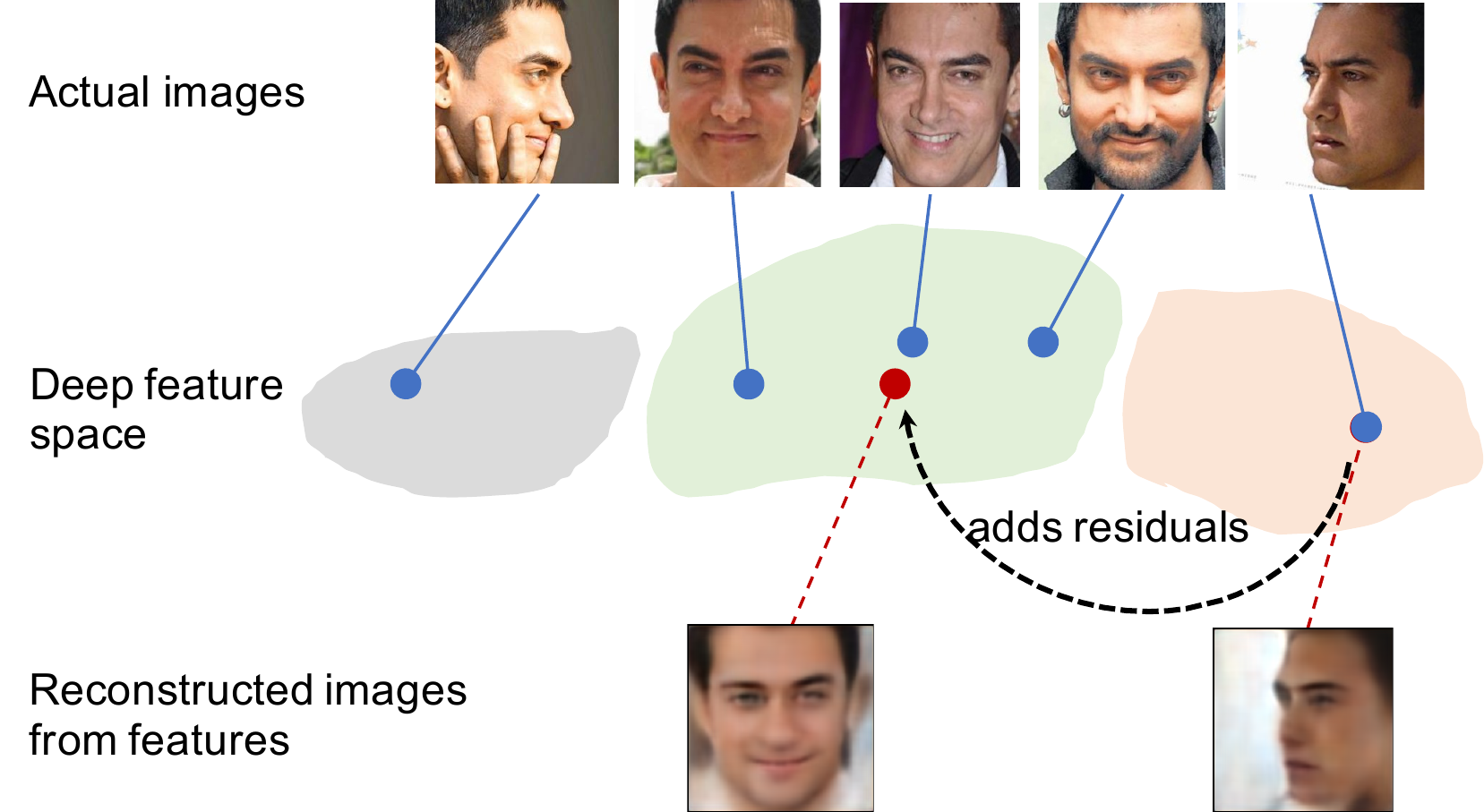}
\caption{At the top of this figure, we illustrate the deep feature embedding of a subject in different poses. The proposed DREAM block is capable of adding residual to the feature of a profile face and map it to the frontal space. At the bottom of the figure, we show the actual reconstructed image of a profile face and its mapped frontal face.}
\label{fig:dream_mapping}
\end{center}
\vspace{-0.25cm}
\vskip -0.5cm
\end{figure}

In this study, we hypothesize that the profile face domain possesses a gradual connection with the frontal face domain in the deep feature space.
Figure~\ref{fig:dream_mapping} illustrates a deep representation embedding of faces belong to the same subject but in different poses.
Given an input image of arbitrary pose, we can actually map its feature to the frontal space through a mapping function that adds residual.
This observation is closely connected to the notion of \textit{feature equivariance}~\cite{lenc2015understanding}, which finds the representation of many deep layers depends upon transformations of the input image. Interestingly, such transformations can be learned by a mapping function from data and the function can be subsequently applied to manipulate the representation of an input image to achieve the desired transformation.

Motivated by this observation, we formulate a novel module called \textit{Deep Residual EquivAriant Mapping} (DREAM) block, which can model the transformation between frontal-profile faces in the high-level deep feature space. The block adaptively adds residuals to an input representation to transform a profile face to a canonical pose to simplify recognition. The residuals are generated conditioned on the preceding feature representation via a few additional weight layers. 
To accommodate input faces of arbitrary pose, a soft gate is introduced to adaptively control the amount of residuals such that more residuals are added to extreme profile faces while keeping the representation unchanged if the input is already in a frontal pose.

Our work is conceptually related to the face frontalization~\cite{tran2017disentangled} in that our approach also performs `frontalization' but not in the image space. We observe from our experiments that transforming profile face features to frontal features could yield better performance than image-level frontalization, which is susceptible to the negative influence of artifacts as a result of image synthesis.
To our best knowledge, this study is the first attempt to perform \textit{profile-to-frontal face transformation in the deep feature space}.

The DREAM block is appealing in several aspects:
\begin{enumerate}[noitemsep, topsep=0pt]
\itemsep0em 
\item It is simple to implement. Specifically, the DREAM block is implemented as a simple yet effective gated residual branch. It can be integrated into existing convolutional neural network (CNN) architectures through stitching the block to the base network. It does not alter the original dimensionality of the face representation and can be trained end-to-end with standard back-propagation. 
\item It is light-weight. It adds only a tiny amount of parameters and computation to the base model.  For instance, it only adds  0.3\% parameters on ResNet-18 and increases its forward time by 1.6\%.
\item The proposed approach helps a base network that already does well in near-frontal face recognition to gain better performance in recognizing faces with extreme pose variation. This is done without elaborated data augmentation and face normalization that is practiced by most existing face recognition studies. For instance, the DREAM block reduces the error of ResNet-18 and ResNet-50 by 16.3\% and 23.7\% on the CFP benchmark~\cite{sengupta2016frontal}. In addition, it also gains 8.5\% improvement on the verification task (TAR@FAR=0.001) and reduces the error rate by 17.5\% on the identification task (Rank-1 accuracy) on the IJB-A dataset~\cite{chen2016unconstrained} for ResNet-18. For ResNet-50, the numbers are 7.0\% and 12.6\%, respectively.
\end{enumerate}

\section{Related Work}
\label{sec:related_work}

\noindent
\textbf{Deep Learning for Face Recognition.} 
Deep learning is the prominent technique for face recognition. Most existing studies deploy CNNs, but with different loss functions, such as contrastive loss~\cite{sun2014deep}, triplet loss~\cite{schroff2015facenet}, and center loss~\cite{wen2016discriminative}. 
Center loss represents the current state-of-the-art approach that learns a center for deep features of each class and simultaneously minimize the distances between the deep features and their corresponding class centers, thus intra-class features variations are reduced and discriminative power of learned features are enhanced.
Prior to center loss, Joint Bayesian~\cite{chen2012bayesian} is widely used to derive a similarity metric for robust face verification. 
The aforementioned metric learning methods facilitate more robust verification given faces with arbitrary poses. Nevertheless, as shown in our experiments, we found that the DREAM block performs better, especially on extreme profile faces (such as those in the Celebrities in Frontal-Profile (CFP) dataset~\cite{sengupta2016frontal}).

\noindent
\textbf{Profile Face Recognition.} It is not new for researchers to take pose variations into consideration \cite{sengupta2016frontal,masi2016pose,zhong2017towards} when dealing with the face recognition problem. 
Existing methods address profile face recognition through elaborated dense 3D facial landmark detection and warping~\cite{taigman2014deepface}, face frontalization~\cite{tran2017disentangled}, or training separate models for learning pose-specific identity features~\cite{masi2016pose}. 
There are alternative approaches.
For instance, Masi \etal~\cite{masi2016we} enhances the performance of CNN through augmenting training and test data with face images differ in 3D shape, expression and pose.
Yin \etal~\cite{yin2017multi} propose a multi-task CNN that exploits side tasks, \eg, pose, to serve as regularizations for learning pose-specific identity features.
%
%
In contrast to the aforementioned studies that require elaborated data augmentation or multi-task training, our approach is light-weight and easy to implement.


\section{Deep Residual Equivariant Mapping}
\label{sec:methodology}

We first present a background on feature equivariance~\cite{lenc2015understanding}, and subsequently use it to motivate the proposed DREAM block. We will then present the block's design and different ways we could employ it to achieve pose-robust face recognition. 

\subsection{Feature Equivariance}
\label{subsec:equivariance}

The notion of feature equivariance is presented in~\cite{lenc2015understanding}.
It looks at how a representation changes upon transformations of the input image. An important finding of this paper is that most of the layers in deep neural networks change in an easily predictable manner with the input. And such transformations can be learned empirically from data.

Formally, a convolutional neural network (CNN) can be regarded as a function $\phi$ that maps an image $\mathbf{x} \in \mathcal{X}$ to a vector $\phi(\mathbf{x})\in \mathbb{R}^d$. The representation $\phi$ is said equivariant with a transformation $g$ of the input image if the transformation can be transferred to the representation output~\cite{lenc2015understanding}. That is, equivariance with $g$ is obtained when there exists a map $M_g : \mathbb{R}^d \rightarrow \mathbb{R}^d$ such that 
\begin{equation}
\label{eqn:equivariance}
 \forall \mathbf{x} \in \mathcal{X}: \;\;\; \phi(g\mathbf{x}) \approx M_g\phi(\mathbf{x}).
\end{equation}
Interestingly, by requiring the same mapping $M_g$ to work for any input image, the function would capture intrinsic geometric properties of the representations.
In~\cite{lenc2015understanding}, the authors focus on geometric transformation such as affine warping and flips of images.
In our problem context, the transformation $g$ is more challenging as it involves 3D geometric changes from profile to frontal faces.

\subsection{Problem Formulation and the DREAM Block}
\label{subsec:formulation}

Face recognition relies on robust representation extracted from deep models. For instance, DeepID series~\cite{sun2014deep} extract features from the penultimate layer (the fully-connected layer before the output layer) as a feature vector and feed it into a classifier. Our goal in this study is to design a light-weight solution to make the representation robust to pose variation.   

We begin by introducing the problem formulation. 
We denote a CNN as a function $\phi$ and the image representation it maps from image $\mathbf{x}$ as $\phi(\mathbf{x})$. We call the network a stem CNN or base network.
Let's assume that we are given two types of face images, namely frontal face image, represented as $\mathbf{x}_f$ and profile face image, denoted by $\mathbf{x}_p$. Note that we assume this setting to facilitate our discussion; the proposed method can work with faces of an arbitrary pose.

Motivated by Eq.~\eqref{eqn:equivariance}, we wish to obtain a transformed representation of a profile image $\mathbf{x}_p$ through a mapping function $M_g$, so that $M_g\phi(\mathbf{x}_p) \approx \phi(\mathbf{x}_f)$.  
To facilitate the incorporation of $M_g\phi(\mathbf{x}_p)$ to a stem CNN, we formulate it as a sum of the original profile feature $\phi(\mathbf{x}_p)$ with residuals given by a residual function $\mathcal{R}(\phi(\mathbf{x}_p))$ weighted by a yaw coefficient $\mathcal{Y}(\mathbf{x}_p)$. That is
\begin{eqnarray}
\label{eqn:residual_mapping}
\phi(g\mathbf{x}_p) & = & M_g\phi(\mathbf{x}_p) \\ \nonumber
                    & = & \phi(\mathbf{x}_p) + \mathcal{Y}(\mathbf{x}_p)\mathcal{R}(\phi(\mathbf{x}_p)) \\ \nonumber
                    & \approx & \phi(\mathbf{x}_f).
\end{eqnarray}
By performing this transformation, we wish that the fixed representation $\phi(\mathbf{x}_p) + \mathcal{Y}(\mathbf{x}_p)\mathcal{R}(\phi(\mathbf{x}_p))$ will be mapped to the frontal face space as illustrated in Figure~\ref{fig:dream_mapping}.

\begin{figure}[t]
\begin{center}
   \includegraphics[width=1\linewidth]{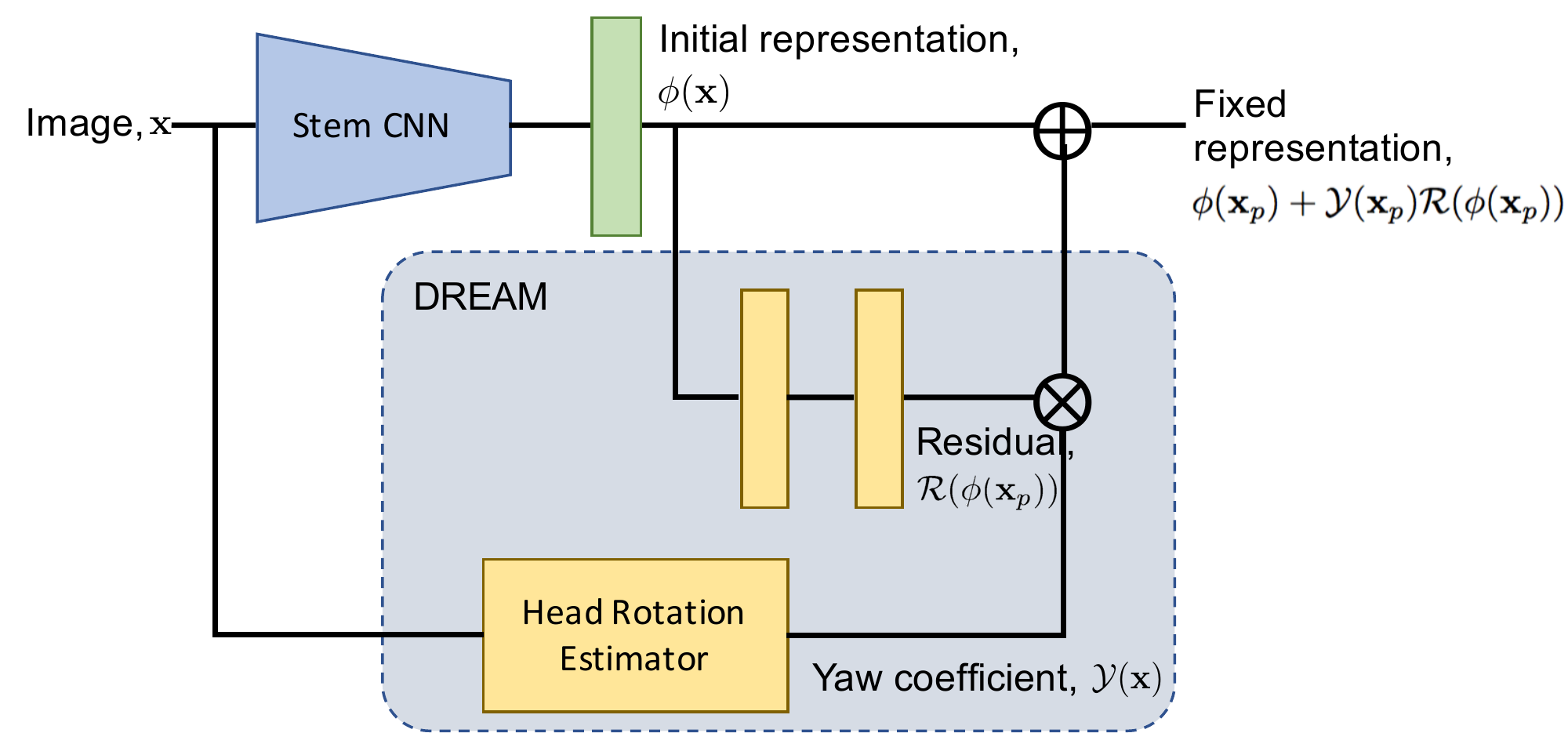}
\end{center}
\vskip -0.4cm
   \caption{The DREAM block is designed with simplicity in mind and it can be easily added to existing CNNs. The block dynamically adds residuals to an input representation to transform a profile face to a canonical pose to simplify recognition.}
\label{fig:architecture}
\vskip -0.2cm
\end{figure}

Equation~\eqref{eqn:residual_mapping} is capable of coping with input images of arbitrary pose, thanks to the yaw coefficient $\mathcal{Y}(\mathbf{x}) \in [0,1]$, which acts as a \textit{soft gate} of the residuals. 
The role of this soft gate is to provide a higher magnitude of residuals (thus a heavier fix) to a face that deviates more from the frontal pose.
Intuitively, $\mathcal{Y}(\mathbf{x})=0$ if the face image is frontal, and its value gradually changes from $0$ to $1$ when the face pose shifts from frontal to a complete profile. The residual's magnitude is thus the largest at the complete profile pose. 
The soft gate is essential. Without it residuals $\mathcal{R}(\phi(\mathbf{x}))$ will be added blindly to input images of any poses, affecting the final face recognition performance.

The soft gated residual block presented in this study is related to the residual structure \cite{he2016deep} that is used to achieve identity mapping in very deep networks.
While \cite{he2016deep} utilizes the residual block to increase the effective depth of networks. Our attempt of combining a soft control gate can be viewed as a correction mechanism that adopts top-down information (the yaw in our case) to influence the feed-forward process. The role of the proposed control gate is to determine the amount of residuals to be passed to the next level.  Besides, it is worth noticing that roll and pitch angles are not considered. The effect of roll will be eliminated by face alignment while face images with large pitch angles are rare and it is possible to address pitch angles by adding another branch in our DREAM block.

\noindent
\textbf{Architecture and Training}.
The residual formulation in Eq.~\eqref{eqn:residual_mapping} allows us to design a succinct network structure shown in Figure~\ref{fig:architecture}. Specifically, we use a stem CNN, \eg, ResNet-18 or ResNet-50, to extract features from the input face image. 
To adapt ResNet~\cite{he2016deep} for our recognition task, we add a 256-dimensional fully connected layer between the average pooling layer and the original fully connected layer. We call this newly added layer as feature layer.
The stem CNN can also be of any of the existing face recognition models~\cite{sun2014deep,schroff2015facenet,wen2016discriminative}.
A fully connected layer is then used to extract the initial representation, $\phi(\mathbf{x})$, which is subsequently `fixed' by the DREAM block. 
The DREAM block, in our current implementation, consists of two branches: 

\vspace{0.1cm}
\noindent
\underline{Residual Branch}. The first branch generates the residuals $\mathcal{R}(\phi(\mathbf{x}))$. It has two fully-connected layers with Parametric Rectified Linear Unit (PReLU)~\cite{he2015delving} as the activation function. 
This branch is learnable separately from the stem CNN. 
Specifically, we train it by minimizing the Euclidean distance between the mapped profile feature and its corresponding frontal feature using stochastic gradient descent.
\begin{equation}
    \min_{\Theta_R} \mathrm{E} \| \phi(\mathbf{x}) + \mathcal{Y}(\mathbf{x})\mathcal{R}(\phi(\mathbf{x})); \Theta_\mathcal{R} ) - \phi(\mathbf{x}_f ) \|^2_2,
\end{equation}
where $\Theta_\mathcal{R}$ denotes the parameters of $\mathcal{R}(\cdot)$. We keep the parameters fixed for the $\mathcal{Y}(\cdot)$ branch.
During the training process we use dropout strategy~\cite{srivastava2014dropout} to the last fully connected layer in the branch. In this work, we train this branch on frontal-profile pairs sampled from MS-Celeb-1M dataset.

\vspace{0.1cm}
\noindent
\underline{Soft Gate with Head Rotation Estimator}. 
The second branch produces the soft yaw coefficient $\mathcal{Y}(\mathbf{x})$. 
This branch assumes an input of 21 facial landmarks following the standard AFLW's protocol~\cite{koestinger2011annotated}.
Note that this requirement does not add additional burden to the stem CNN since the face alignment process is a standard preprocessing step of many face recognition pipelines\footnote{In many face recognition pipelines, face alignment is used to center the face, rotate the face such that the eyes lie along a horizontal line, and scale the faces such that they are approximately identical in size.}. 

Given facial landmarks, the head rotation estimator in the second branch (see Figure~\ref{fig:architecture}) estimates the head rotation by using the algorithm presented in~\cite{zhang2015appearance}. Specifically, we adopt the same definition of the face model and head coordinate system as~\cite{sugano2014learning}. Slightly different from~\cite{zhang2015appearance}, we extend their 6-point 3D face model into a 21-point model to achieve a better performance. We then fit the model by estimating the initial solution using the EPnP algorithm~\cite{lepetit2009epnp}, and further refining the pose via non-linear optimization.

The yaw angle obtained from previous steps is then non-linearly mapped to a positive value within the range of $[0,1]$.
Specifically, we obtain the training target of yaw coefficient by $ \sigma( \frac{4}{\pi}y - 1 )$, where $y$ is the yaw angle of a face image in radian units. The $\sigma$ here is a sigmoid function. Through this mapping the coefficient quickly reaches the value 1 once a face turns more than 45$^\circ$, exerting more residuals for extreme profile faces. 
Empirically, we found that by adding a monotonous nonlinear mapping to the coefficient improves the effectiveness of residuals to the representation. 

The outputs of these two branches are multiplied and added to the initial representation $\phi(\mathbf{x})$. The resulting feature $\phi(\mathbf{x}) + \mathcal{Y}(\mathbf{x})\mathcal{R}(\phi(\mathbf{x}))$ is the final feature output.
It is worth noting that pose estimation accuracy has little effect on the final performance. We use the state-of-the-art face alignment method proposed in \cite{zhu2016unconstrained} to obtain the 21 facial landmarks. The method is specially designed to handle large-pose alignment. We have tried adding 20\% noise to the estimated yaws in both training and test sets and found that EER in face recognition increases by no more than 2\% under multiple settings. This result is still a lot more better than baselines.

\subsection{The Usages of DREAM}
\label{subsec:strategy}

We describe three ways of utilizing DREAM. Comparative results on all strategies will be provided in the experiment section.

\noindent
\textbf{Stitching}.
The most convenient way of deploying the DREAM block is by `stitching' the block directly to a trained stem CNN. In particular, given a base network, we can just stitch the DREAM block onto the final feature layer of the base network without changing any learned parameters of the original model.

\noindent
\textbf{End-to-end}.
The proposed light-weight block can also be trained together with the stem CNN in an end-to-end manner. Given a plain base network, we insert the DREAM block and directly train the new network with random initialization on all of its parameters. 
If the stem CNN is not plain but trained previously, we can first fine-tune the stem CNN while training the DREAM block end-to-end using an existing face recognition loss (\eg, verification loss, identification loss, or both). We name the strategy as `end2end'. With this strategy, the performance on profile faces is not guaranteed since the DREAM block may not be able to distinguish frontal and profile cases, since no specific frontal-profile face pairs are used for training the block.

\noindent
\textbf{End-to-end$+$retrain}.
We train the stem CNN and DREAM block together then train the DREAM block separately with frontal-profile face pairs.

\begin{figure}[t]
\begin{center}
   \includegraphics[width=1\linewidth]{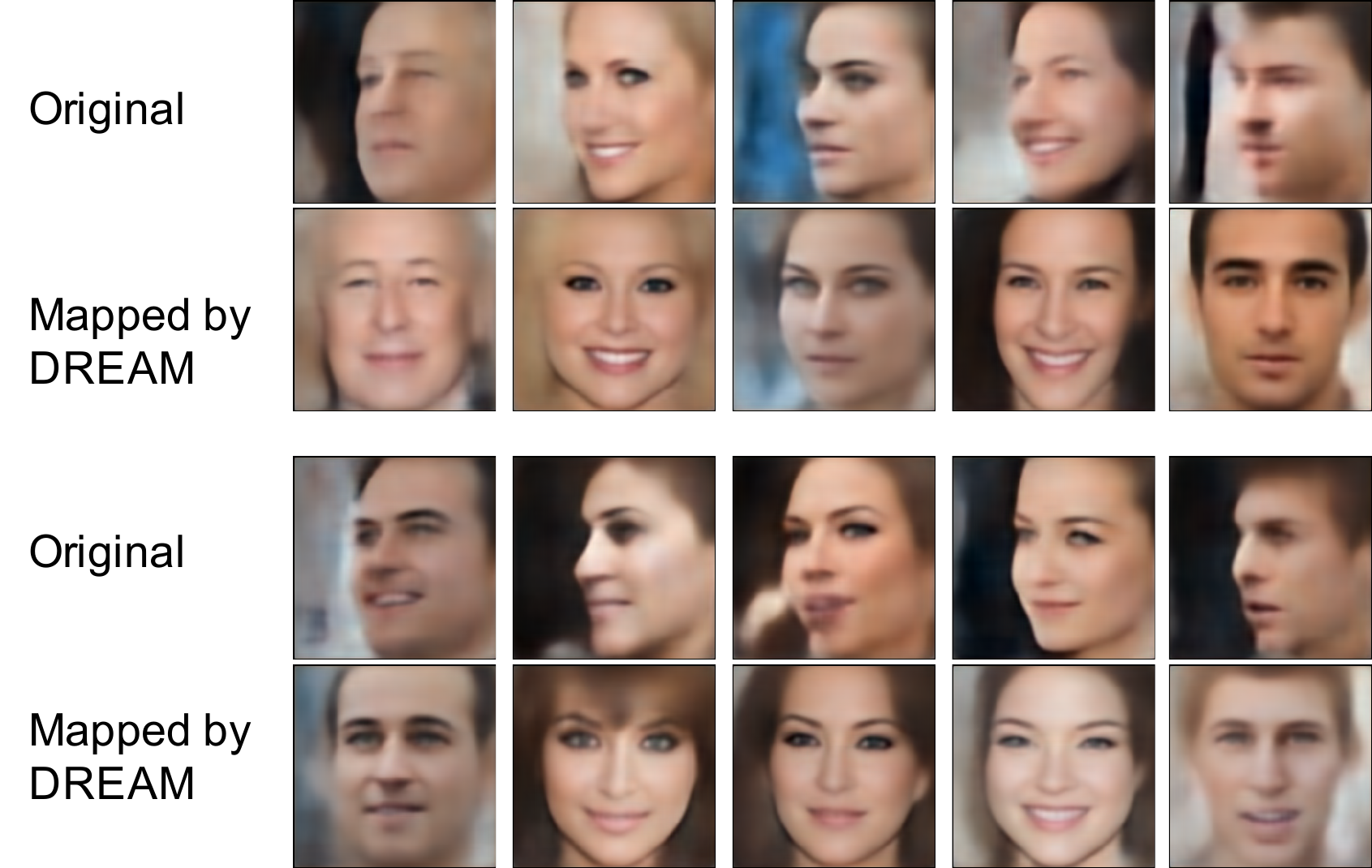}
\end{center}
\vskip -0.2cm
   \caption{Visualization of deep features. The first and third rows show the reconstructed original features of profile faces. The second and fourth row depict the reconstructed features after the mapping by DREAM block.}
\label{fig:transform}
\end{figure}

As a way to demonstrate the effectiveness of DREAM block, we use the GAN model in \cite{nguyen2016plug} that could map deep features back to reconstructed images. This GAN model is used to visualize the original and mapped features generated by our model. Some representative results are shown in Fig.~\ref{fig:transform}. Note that the reconstruction only serves for visualization purposes. The usefulness of DREAM can only be fully validated through examining its performance on the face recognition task. We show strong performance of DREAM in the following section.
 
\vskip -0.2cm
\noindent

\section{Experiments}
\label{sec:experiments}



We divide our experiments into a few sections. In the first section, we show the effectiveness of our method on the task of frontal to complete profile face verification.
In the second section, we compare with state-of-the-art methods on a dataset with full-pose variation. Lastly, we provide a further analysis of the influence of face yaw and present an ablation study.

\begin{table*}[t]
\footnotesize
\begin{center}
\caption{Results on Celebrities in Frontal-Profile (CFP) with Frontal-Profile setting. Equal error rate (EER) is reported. Lower is better. The best result in each row are given in bold.}
\vskip -0.1cm
\resizebox{2.1\columnwidth}{!}{%
\begin{tabular}{ll|c|ccc|ccc}
\hline
 &  &  & \multicolumn{3}{|c|}{Other Strategies} & \multicolumn{3}{|c}{DREAM Variants} \\
Model & Training Data & Na\"{i}ve & CDFE \cite{lin2006inter} & JB \cite{chen2012bayesian} & FF~\cite{tran2017disentangled} & stitching & end2end & end2end+retrain \\
  \hline \hline
  ResNet-18 & MS-Celeb-1M & 8.40  & 8.30 & 8.37 & 14.40 & 7.71 & 7.63 & \textbf{7.03} \\
  ResNet-50 & MS-Celeb-1M & 7.89 &  7.71   &  8.49    & 14.26   &  7.29    &  6.43    & \textbf{6.02}     \\     
  Center-Loss & MS-Celeb-1M & 8.54  & 8.49 & 8.29 & 14.53 & 7.82 & 7.81 & \textbf{7.26}  \\ \hline
\end{tabular}}
\label{tab:mainExperiment}
\end{center}
\vspace{-0.25cm}
\vskip -0.4cm
\end{table*}

We provide a description of the training set we use before moving to the next section.
To train our own stem CNNs (ResNet-18 and ResNet-50) and DREAM block, we employ a subset of MS-Celeb-1M~\cite{guo2016ms}. The original data provides images of 100,000 top celebrities in the form of URL, which, however, are collected from Google without manual cleaning. To facilitate the learning of our model, we clean the data and select a subset to form our training and test partitions. The training partition consists of 696,446 images from 13,385 identities. The testing partition contains 70,730 images from 3,084 celebrities. The training and test partitions have exclusive identities. The test partition is only used in our further analysis and ablation study since it contains more profile faces. 
All face images in training and testing sets are prepocessed to a size of $224\times224$. We use the method presented in \cite{zhang2016joint} and \cite{zhu2016unconstrained} for face detection and face alignment, respectively.
To train our stem CNN, we use an identification loss~\cite{sun2014deep}. As in~\cite{nguyen2010cosine}, face verification is performed by measuring the cosine distance between feature representation of queries.

\subsection{Evaluation on CFP with Frontal-Profile Setting}

\noindent
\textbf{Test dataset}.
We first conduct evaluations on the Celebrities in Frontal-Profile (CFP) dataset~\cite{sengupta2016frontal}, a challenging dataset created to examine the problem of frontal to profile face verification `in the wild'. 
The dataset contains 500 celebrities, each of which has ten frontal and four profile face images. We follow the standard 10-fold protocol~\cite{sengupta2016frontal} in our evaluation. In particular, the whole dataset is divided into 10 folds each containing 350 same and 350 not-same pairs generated from 50 individuals (7 same and 7 not-same pairs for each individual). The same protocol is applied on both the Frontal-Profile and Frontal-Frontal settings.

\noindent
\textbf{Stem CNN}.
To show the benefits of DREAM on different base networks, we perform our experiments using the following stem CNN architectures. Following \cite{wu2016deep}, we employ a ResNet-18 as our base network. We also try ResNet-50.  
In addition, we experiment with a state-of-the-art network presented in \cite{wen2016discriminative}\footnote{We use the codes released by the authors at \url{https://github.com/ydwen/caffe-face}.}, which stacks 11 residual blocks and trains with Center-Loss. 

\noindent
\textbf{Baselines}.
We compare our method with three representative baselines that also help to alleviate the gap in frontal-profile face verification:
\begin{itemize}[noitemsep, topsep=0pt]
\itemsep0em 
\item CDFE~\cite{lin2006inter} -  Common Discriminant Feature Extraction is a representative method that is specially tailored to the inter-modality problem. In the algorithm, two transforms are simultaneously learned to map the samples in two modalities respectively to the common feature space.
\item JB~\cite{chen2012bayesian,cao2013practical} - Joint Bayesian is a widely used metric learning approach in face verification, which also supports the transfer from one domain to another. We train a JB model on the CFP dataset. Each time, we use 9 splits for training and the rest for testing. We repeat this process for 10 times and report the average error. The DREAM block does not need such fine-tuning.
\item FF~\cite{tran2017disentangled} - Face Frontalization morphs faces from profile to frontal with a generative adversarial network. This baseline converts all profile face images in the CFP test partition to frontal ones.
\end{itemize}
In addition, we also evaluate the effectiveness of different training strategies of DREAM as discussed in Section~\ref{subsec:strategy}, namely, `stitching', `end2end', and `end2end$+$retrain'.

\vspace{0.1cm}
\noindent
\textbf{Results}.
The results are summarized in Table~\ref{tab:mainExperiment}. We report the Equal Error Rate (EER). 
We have following observations:

\begin{figure}[t]
\begin{center}
   \includegraphics[width=0.85\linewidth]{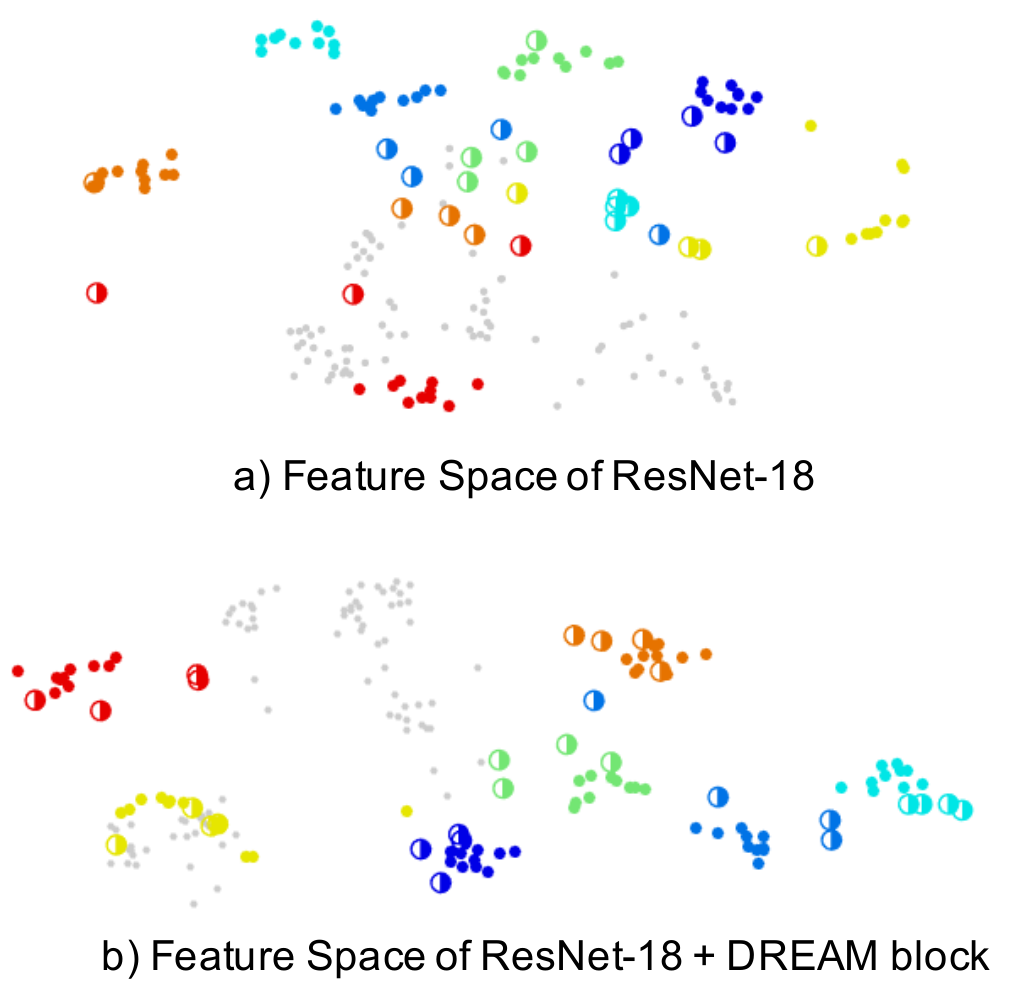}
\end{center}
\vskip -0.4cm
   \caption{Visualization of deep feature space. Here we use solid dots $\bullet$ to represent frontal faces and symbols \RIGHTcircle~to denote profile faces. The features of different subjects are represented in different colors. Figure (a) shows the embedding yielded by a ResNet-18 model. It tends to cluster profile faces around the same point in the space. Figure (b) shows the embedding obtained by ResNet-18 with the DREAM block. Profile faces are separated more clearly and clustered with their own class of frontal faces.}
\label{fig:tSNE}
\vskip -0.25cm
\end{figure} 

\noindent
1) Through the comparisons between Na\"{i}ve and DREAM variants, we found that all DREAM strategies are capable of reducing the EER of various strong models, in spite of their different designs and training sources. The strategy `end2end$+$retrain' gives the best performance overall. Through visualizing the deep feature space of `Na\"{i}ve' ResNet-18 via t-SNE in Figure~\ref{fig:tSNE}, we could observe that features of frontal faces are clearly separated based on identities, but the profile ones are mostly messed up exhibiting severe feature overlapping in the space. In contrast, ResNet-18 with the proposed DREAM block clearly separates the features of different subjects without affected by the pose factor.

\noindent
2) Baselines such as CDFE, JB and FF utilize the same supervision as our approach but they are less effective. The fact that FF performs poorer than DREAM in this case suggests that performing `frontalization' in the deep feature space rather than the image space turns out to be more fruitful. We conjecture that the artifacts of `frontalized' faces lead to the inferior results of FF to Na\"{i}ve baseline. Some frontal images synthesized by FF are given in the supplementary material.

\noindent
3) Our best result, which is obtained by ResNet-18 and ResNet-50 trained with MS-Celeb-1M and `end2end+retrain' strategy, outperforms many existing methods, including \cite{chen2016fisher}, \cite{sankaranarayanan2016triplet}, and \cite{sengupta2016frontal} that achieves 8.00, 8.85, and 14.97, respectively.

It is worth pointing out that the impact introduced by the proposed block on the frontal face recognition is minimal. In particular, the block hardly affects the performance of non-profile face recognition since the soft yaw coefficient prevents the block from altering the frontal representation. In particular, for ResNet-18, the EER of `Na\"{i}ve' on Frontal-Frontal setting is 2.66, while that of `stitching', `end2end'. and `end2end+retrain' are 2.69, 2.00 and 2.00, respectively.

\subsection{Evaluation on IJB-A with Full Pose Variation}

\begin{table*}[t]
\footnotesize
\begin{center}
\caption{Comparative performance analysis on IJB-A benchmark. Results reported are the `average$\pm$standard deviation' over the 10 folds specified in the IJB-A protocol. Symbol `-' indicates that the metric is not available for that protocol. Standard deviation is not available for all the methods. f.t. denotes fine tuning a deep network multiple times for each training split.}
\vskip -0.2cm
\resizebox{2.1\columnwidth}{!}{
\begin{tabular}{l|c|c|c|c}
 \hline
 \multirow{1}{*}{Methods $\downarrow$} & \multicolumn{2}{c|}{Verification} & \multicolumn{2}{c}{\multirow{1}{*}{Identification}} \\
 \hline
 \multirow{1}{*}{Metrics $\rightarrow$} & TAR @ FAR=0.01 & TAR @ FAR=0.001 & Rec. Rate @ Rank-1 & Rec. Rate @ Rank-5 \\
 \hline \hline
  \textbf{Our Approach with MS-Celeb-1M subset:} &  & & &  \\ 
  ResNet-18 (na\"{i}ve) &  0.840$\pm$0.026 & 0.656$\pm$0.040 & 0.897$\pm$0.016 & 0.951$\pm$0.011 \\ 
  ResNet-18 (end2end$+$retrain) &    0.872$\pm$0.018 & 0.712$\pm$0.035 & 0.915$\pm$0.012  & 0.962$\pm$0.008 \\
  ResNet-50 (na\"{i}ve) &  0.881$\pm$0.018 & 0.714$\pm$0.034 & 0.913$\pm$0.013 & 0.957$\pm$0.010 \\ 
  ResNet-50 (end2end$+$retrain) &    0.891$\pm$0.016 & 0.764$\pm$0.031 & 0.924$\pm$0.016  & 0.962$\pm$0.010 \\
  \hline
  \textbf{Our Approach with full MS-Celeb-1M:} &  & & &  \\ 
  ResNet-18 (na\"{i}ve)& 0.934$\pm$0.009 & 0.836$\pm$0.016 & 0.939$\pm$0.012 & 0.960$\pm$0.010 \\
  ResNet-18 (end2end$+$retrain)& \textbf{0.944$\pm$0.009} & 0.868$\pm$0.015 & \textbf{0.946$\pm$0.011}  & \textbf{0.968$\pm$0.010} \\

\hline
  \textbf{Existing Methods:} &  & & &  \\ 
  Wang \etal \cite{wang2017face} & 0.729$\pm$0.035 & 0.510$\pm$0.061& 0.822$\pm$0.023 & 0.931$\pm$0.014 \\
  Pooling Faces \cite{hassner2016pooling} & 0.819$\pm$ ------ & 0.631$\pm$ ------ & 0.846$\pm$ ------ & 0.933$\pm$ ------ \\ 
  Deep Multi-Pose \cite{abdalmageed2016face} & 0.787$\pm$ ------ & -- & 0.846$\pm$ ------ & 0.927$\pm$ ------ \\
  Multi-task CNN ~\cite{yin2017multi} & 0.787$\pm$0.043  & -- & 0.858$\pm$0.014 & 0.938$\pm$0.009 \\
  PAMs \cite{masi2016pose} & 0.826$\pm$0.018 & 0.652$\pm$0.037 & 0.840$\pm$0.012 & 0.925$\pm$0.008 \\   
  DCNN$_\mathrm{fusion}$ (f.t.) \cite{chen2016unconstrained} &  0.838$\pm$0.042 & -- & 0.903$\pm$0.012  & 0.965$\pm$0.008 \\
  Augmentation$+$Video Pooling$+$Rendered Test \cite{masi2016we} & 0.886$\pm$ 0.017 & 0.725$\pm$ 0.044 & 0.906$\pm$ 0.013 & 0.962$\pm$ 0.007 \\ 
  CNN$_\mathrm{media}$+TPE (f.t.)~\cite{sankaranarayanan2017triplet} & 0.900$\pm$0.010 & 0.813$\pm$0.020 & 0.932$\pm$0.010 & -- \\
  Template Adaptation (f.t.)~\cite{crosswhite2017template} & 0.939$\pm$0.013 & -- & 0.928$\pm$0.010 & -- \\  
  Quality Aware Network (f.t.)~\cite{liu2017quality} & 0.942$\pm$0.015 & \textbf{0.893$\pm$0.039} & -- & -- \\  
\hline
 \end{tabular}
 }
 \label{tab:IJBA}
\end{center}
\vskip -0.4cm
\end{table*}

\noindent
\textbf{Test dataset}.
In previous experiments on CFP we focus on frontal-profile face verification. In this experiment, we further evaluate our method on another challenging benchmark called IARPA Janus Benchmark A (IJB-A)~\cite{klare2015pushing} that covers full pose variation (yaw angles between $-90^\circ$ to $+90^\circ$). 
The dataset contains 500 subjects with of 5,712 images and 20,414 frames extracted from videos. The faces in the IJB-A dataset contain extreme poses and illuminations, more challenging than LFW \cite{huang2007labeled}. 
Following the standard protocol in~\cite{klare2015pushing}, we evaluate our method on both verification (1:1) and identification tasks (1:N).

\noindent
\textbf{Stem CNN}.
We use ResNet-18 and ResNet-50 trained on MS-Celeb-1M as our stem CNN. The DREAM block is deployed using our `end2end$+$retrain' strategy.

\noindent
\textbf{Results}.
Table \ref{tab:IJBA} reports our results on IJB-A. It is noteworthy that we employ a strong baseline: a ResNet-18 trained on the MS-Celeb-1M dataset, which achieved 65.6\% True Acceptance Rate (TAR) at False Acceptance Rate (FAR) of 0.001 on the verification task and a Rank-1 recognition accuracy of 89.7\% on the identification task, comparable to a state-of-the-art method~\cite{chen2016unconstrained}.
%
By adding the DREAM block, the performance of this strong baseline is improved by \textbf{8.5}\% on the verification task (TAR@FAR=0.001) and the error rate is reduced by \textbf{17.5}\% on the identification task (Rank-1 accuracy). For the even stronger ResNet-50, the DREAM block also offers a compelling error reduction of 7.0\% and 12.6\%, respectively.

The aforementioned results are generated by using a sampled subset of MS-Celeb-1M dataset for training. Once we use the full MS-Celeb-1M dataset, we achieve the state-of-the-art results on the identification task and comparable verification performance against Quality Aware Network~\cite{liu2017quality} that uses a more elaborated loss (triplet$+$softmax) for training.

There are other strategies that can be used to further improve the performance but we did not attempt each of them.
We included these methods in Table~\ref{tab:IJBA} for completeness.
Most of the techniques perform excellently when they fine-tune their model multiple times for each training split. Such methods are denoted with `f.t.' in Table \ref{tab:IJBA}. Our model is trained on MS-Celeb-1M and directly tested on IJB-A without fine-tuning.
Data augmentation technique can be useful too. Masi~\etal~\cite{masi2016we} perform task-specific data augmentation on pose, shape, and expression on CASIA-WebFace. Their method also performs pose synthesis at test time. We believe the same technique can be used to improve the performance of our approach.
Sankaranarayanan \etal~\cite{sankaranarayanan2017triplet} deploy a metric learning approach to fine-tune the network on the training splits of IJB-A. Liu \etal~\cite{liu2017quality} train a network with joint triplet and softmax losses.
Our method is only trained using the typical identification loss and without fine-tuning.

\begin{figure}[t]
\centering
   \includegraphics[width=0.95\linewidth]{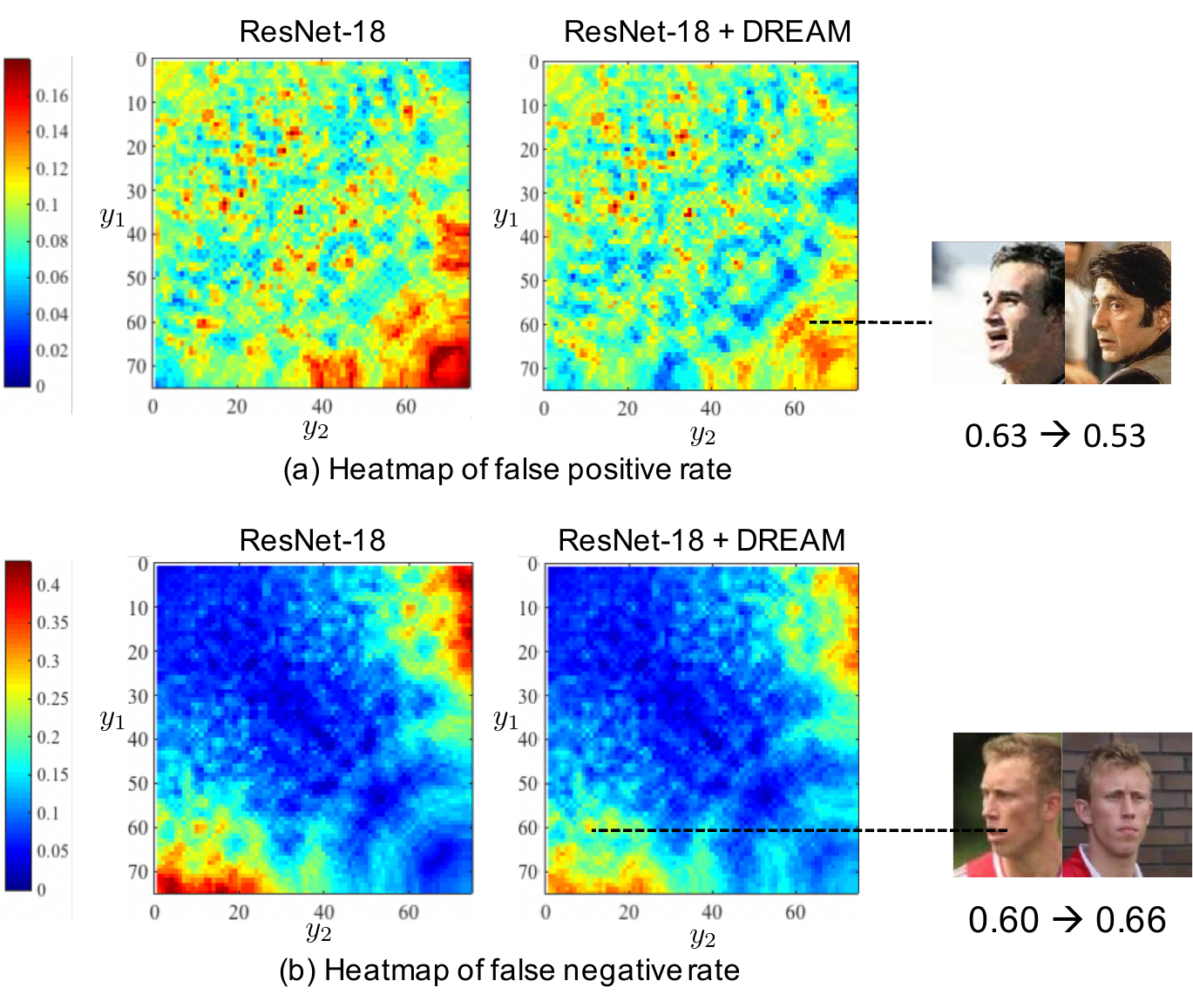}
   \caption{A comparison between the false positive rate and false negative rate between the Na\"{i}ve ResNet-18 and ResNet-18 $+$ DREAM on the yaw space. (a) and (b) show heatmaps of False positive rate and False negative rate given face pairs of different yaws $(y_1,y_2)$.
   $0.63 \rightarrow 0.53$ means the cosine similarity drops from $0.63$ to $0.53$ after the DREAM block was utilized.}
\label{fig:yawAnalysis}
\vskip -0.4cm
\end{figure}

\subsection{A Further Analysis on Influences of Face Yaw}

We conduct a more in-depth analysis on the influence of face yaw to the performance of face verification to better understand the effectiveness of DREAM block on profile face recognition. 
In the analysis, we assume a query face and a matching face with yaw angle $(y_1,y_2)$, respectively. We then select a threshold of equal error rate. Given image pairs of different $(y_1,y_2)$ and the threshold, we quantify their false positive rate and false negative rate and plot these values as heatmaps in Figure~\ref{fig:yawAnalysis}. We compare the performance of two approaches, namely the Na\"{i}ve ResNet-18 as a baseline, and a ResNet-18 equipped with the proposed DREAM block. 
It can be observed from Figure~\ref{fig:yawAnalysis}(a) that false positives mainly concentrate at the bottom right of the heatmap, \ie, when both query and matching faces are of extreme yaw angles. In comparison to the baseline, our approach yields much fewer false positives. 
From Figure~\ref{fig:yawAnalysis}(b), we observe that a majority of false negatives take place when one of the faces is frontal and the other one is profile. Again, our approach performs superior over the baseline in terms of false negative rates.

\subsection{Examining the Architecture of DREAM Block}

In this subsection we study the effectiveness of different architectures of the DREAM block. The architecture of DREAM block could differ in two aspects: 1) the location to apply the block, and 2) the design. For the first part, we compare different locations of inserting a DREAM block.
`After pooling layer' means we insert the DREAM block directly after the average pooling layer of ResNet. 
Recall that we introduced a new fully connected layer between the average pooling layer and the original fully connected layer of ResNet (see Sec.~\ref{subsec:formulation}). `After feature layer' refers to inserting the DREAM block after this feature layer.
%
For the second part, we try to build the DREAM block with only one fully connected layer or replace one fully connected layer with a 1D convolution layer. 
We found that the best location to insert DREAM block is at the top of the network, where the feature is deep and feature dimension is compact. We believe the two aspects will benefit the learning of DREAM block. From Table \ref{tab:dreamBlockStructure} we observe that non-linearity plays a role in DREAM block (the setting that uses 2 fc layers outperforms linear mapping).

\begin{table}[t]
\footnotesize
\caption{Comparative analysis of different architectures of DREAM block. Evaluation is conducted on MS-Celeb-1M with ResNet-18. Results are reported as EER error.}
\vskip 0.1cm
\begin{adjustbox}{width=1\linewidth}
\begin{tabular}{c|c|c}
   \hline
  Position of DREAM &  Components of DREAM & EER on MS-Celeb-1M \\
  \hline\hline
  After pooling layer &  two fc & 9.22   \\
  After feature layer &    one fc     &   8.92      \\ 
  After feature layer & one fc, one conv  & 9.18      \\
  After feature layer & two fc      & \textbf{8.45}  \\
  \hline
 \end{tabular}
 \end{adjustbox}
\label{tab:dreamBlockStructure}
\vskip -0.1cm
\end{table}

\subsection{An Ablation Study on the Soft Gate}

The soft gate produces a yaw coefficient to control the amount of residuals to be added to the initial representation. As described in the methodology section, the coefficient can be linear or nonlinear w.r.t. the degree of face yaw. In this experiment, we examine the effectiveness of different settings of the soft gate. 
In addition, we also report a baseline in which the soft gate is consistently closed, \eg, $\mathcal{Y}(\mathbf{x})=1$. It is noteworthy that this baseline still keeps the same number of parameters but it degenerates to a conventional residual branch that losses the capability of distinguishing frontal and profile faces.

\begin{table}[t]
\footnotesize
\vskip 0.1cm
\centering
\caption{Comparative analysis of different gate settings. Evaluation is conducted on MS-Celeb-1M with ResNet-18. Results are reported as EER error.}
\vskip 0.1cm
\begin{tabular}{c|c}
   \hline
  Gate Setting &  With DREAM Block  \\
  \hline\hline
  Consistently Close, $\mathcal{Y}(\mathbf{x})=1$ & 9.31      \\
  Linear & 8.82      \\ 
  Nonlinear & \textbf{8.45}      \\ 
  \hline
 \end{tabular}
\label{tab:celebReserve}
\vskip -0.2cm
\end{table}

From Table \ref{tab:celebReserve}, we observe that the performance of face verification significantly drops if the soft gate is closed consistently.
The results suggest the improved performance attained by the DREAM block insertion is not merely due to the additional parameters, but the effective mechanism inherently brought by the residual branch and soft gate.
From Table \ref{tab:celebReserve}, we also observe that nonlinear mapping yields superior performance over the linear setting. This observation ascertains our design of exerting a higher degree of correction to a face pose larger than 45$^\circ$, as this pose range is harder to be handled by the stem CNN (as observed from Figure~\ref{fig:yawAnalysis}).

\section{Conclusion}

We have presented a Deep Residual EquivAriant Mapping (DREAM) block to improve the performance of face recognition on profile faces. 
Our method is novel in that we take a radically different approach to handle profile faces. 
Specifically, we bridge the discrepancy between profile and frontal faces through performing equivariant mapping in the deep feature space. The mapping is achieved through the light-weight DREAM block that is easy to implement. Extensive results on CFP, IJB-A, and MS-Celeb-1M datasets demonstrate the applicability of the block on different types of stem CNNs, including ResNet-18, ResNet-50, and Center-Loss models. Interestingly, we observed that performing frontalization in the feature space is more fruitful than the image space for the task of face verification.
It is noteworthy that the proposed block is not limited to face recognition with pose variation, it is suitable for other problems, \eg~cross-age face recognition, of which performance also suffers from uneven distribution of training data. 

\vspace{0.15cm}
\noindent \textbf{Acknowledgement:} This work is supported by SenseTime Group Limited and the General Research Fund sponsored by the Research Grants Council of the Hong Kong SAR (CUHK 14241716, 14224316. 14209217).

\newpage

{\small
\bibliographystyle{ieee}
\bibliography{short,dream}
}

\end{document}